\documentclass[runningheads]{llncs}
\usepackage[T1]{fontenc}
\usepackage{graphicx}
\usepackage{tabularx}
\usepackage{float}
\usepackage{amsmath}
\usepackage{adjustbox}
\usepackage[gen]{eurosym}
\usepackage{subfigure}

\begin{document}
\title{Reducing the Price of Stable Cable Stayed Bridges with CMA-ES}

%

\author{Gabriel Fernandes \inst{1}\orcidID{0000-0001-6662-8675} \and Nuno Lourenço\inst{1}\orcidID{0000-0002-2154-0642}  \and Jo\~ao Correia\inst{1}\orcidID{0000-0001-5562-1996} }
\institute{University of Coimbra, CISUC, DEI, Coimbra, Portugal \texttt{gabrielf@student.dei.uc.pt, naml@dei.uc.pt, jncor@dei.uc.pt}}


%
\authorrunning{G. Fernandes et al.}
%

%
\maketitle              
\begin{abstract}
The design of cable-stayed bridges requires the determination of several design variables' values. Civil engineers usually perform this task by hand as an iteration of steps that stops when the engineer is happy with both the cost and maintaining the structural constraints of the solution. The problem's difficulty arises from the fact that changing a variable may affect other variables, meaning that they are not independent, suggesting that we are facing a deceptive landscape. 

In this work, we compare two approaches to a baseline solution: a Genetic Algorithm and a CMA-ES algorithm. There are two objectives when designing the bridges: minimizing the cost and maintaining the structural constraints in acceptable values to be considered safe. These are conflicting objectives, meaning that decreasing the cost often results in a bridge that is not structurally safe. The results suggest that CMA-ES is a better option for finding good solutions in the search space, beating the baseline with the same amount of evaluations, while the Genetic Algorithm could not. In concrete, the CMA-ES approach is able to design bridges that are cheaper and structurally safe. 


\keywords{Genetic Algorithm  \and CMA-ES \and Optimization \and Cable Stayed Bridges}
\end{abstract}
\section{Introduction}

Bridges are critical components of every transportation network infrastructure. They must be designed to be safe, robust and durable and simultaneously cost-effective and, sometimes, aesthetic pleasing, which are often competing objectives \cite{h1997,Hibbeler2015,Zienkiewicz2014,Chen2014}.
The restrictions on the structural design standards vary from country to country and specify the requirements that bridges must satisfy, such as safety versus heavy vehicle loads, high-velocity winds and earthquakes. The bridge must also be within serviceability requirements which specify the maximum deflections, stresses and oscillations when subject to dynamic actions, such as pedestrians' movements.
Each bridge is planned by a structural design firm or a consortium of several companies. The process usually starts with a tender for the services, in which the choice of the company is based on a set of evaluation criteria where, usually, the best commercial proposal (the lowest price) wins. Design firms must deliver a solution using the lowest possible resources (man-hours).
For this reason, structural designers do not have the time to evaluate all possible solutions, even with simpler designs. As such, any technique or mechanism to help automate and optimize the design of bridges is, therefore very valuable as even a small percentage of optimization (without compromising the bridge safety and requirements) constitutes large sums of money saved for the public treasury.

Cable-stayed bridges (CSB) are one of the most complex type of bridges to design, due to the fact that they are highly static indeterminate. Thanks to the progress that we have seen in computational technologies, we can now build CSBs that are longer but, at the same time, safer.

In this work, we extend the study conducted in \cite{DBLP:conf/gecco/CorreiaFM20,DBLP:conf/evoW/CorreiaF20} which uses an Evolutionary Computation based approach to tackle the problem of designing CSBs. In concrete, the authors propose the use of a standard Genetic Algorithm (GA)\cite{Mitchell:1996} using a representation based on real numbers to represent each parameter that one most optimise to design a bridge. The results attained by the proposed approach were encouraging since the GA was able to optimize this type of bridge (with some variables fixed) in terms of the structural constraints. However, it was not able to reduce the costs when compared to a hand design without resorting to some tuning. For this work, we use the Covariance Matrix Adaptation Evolution Strategy algorithm (CMA-ES)\cite{DBLP:journals/ec/HansenO01} to see if it is able to surpass the results attained by the GA with the same amount of evaluations as well as the baseline solution.


In terms of contributions, we enumerate the following: (i) a study of a more complex problem, due to the number of cables being also evolved, instead of being static; (ii) a comparison between two optimization algorithms, GA and CMA-ES; (iii) the results suggest that a standard GA might not be enough to find efficient solutions.

Additionally, the CMA-ES algorithm was able to discover a bridge with a cost that is 4.656k \euro{} less than the one of the baseline solution. Taking into account that the solution was discovered automatically, without any human input, the result is impressive and opens for further application of evolutionary approaches in the automatic design of bridges.

The remainder of the paper is divided as follows. Section 2 briefly presents the related work. In Section 3, the problem is defined and in Section 4, it is explained how the GA and CMA-ES experiments were modelled. The obtained results are shown in section 5, and our conclusions are listed in Section 6.

\section{Related Work}

The first works on the optimum design of CSBs focused on addressing the cable tensioning problem with fixed geometry and structural sections \cite{Qin1992,Sung2006,Baldomir2010}. More recently, Genetic Algorithms (GA) have also been used to tackle this problem \cite{Hassan2010,Hassan2013}.

Including dynamic loads creates additional constraints for the design problem. Previous researches have focused on earthquakes \cite{Simoes1999,Ferreira2011}, wind aerodynamics \cite{Baldomir2013,Jurado2008,Nieto2011} and pedestrian induced action in cable stayed footbridges \cite{Ferreira2012,Ferreira2019,Ferreira2019a}. These dynamic loads may cause the bridge to vibrate, which is something that is detrimental. To mitigate this problem, there are some options, such as: (i) improve the sturdiness of the bridge by increasing its mass. This is something that is not desirable due to the potential increase in cost and the possibility of the resultant bridge not being as aesthetically pleasing as one might want; (ii) Including control devices like the ones used to retrofit the London Millennium Footbridge, \cite{Dallard2001,dallard2001london}, for example, viscous dampers or tuned mass dampers (TMDs).

Gradient-based optimization techniques have been used to optimize the bridge's geometry, sizing and cable tensioning \cite{Negrao1997,Simoes2000}. GAs have also been used to optimize simultaneously these bridge properties \cite{Hassan2015}, although with simpler models than the ones found in this and in the works on which this article is based \cite{DBLP:conf/gecco/CorreiaFM20,DBLP:conf/evoW/CorreiaF20}. These are based on the works of Ferreira and Simões \cite{Ferreira2019,Ferreira2019a}, from where an already optimized solution (not necessarily a global optimum) was retrieved and then used as a baseline to help us understand if a given solution is in fact good. The literature tells us that gradient-based approaches are able to achieve more rapidly good results than the GA ones. However, compared to a GA, a gradient-based solution is far harder to parameterize, requiring more time and effort, while a GA is easier to get up and running.

As far as we know, there are no works applying CMA-ES to CSB design optimization. The CMA-ES ability to explore the search space via the exploitation of co-variance matrix properties of the genotype holds the potential to further optimize CSBs, beyond existing GA approaches.

\section{Problem Definition}
In this work, we are evolving configurations for cable-stayed footbridges, minimizing the overall cost of the structure while guaranteeing its structural safety. The structural safety of the bridge is accomplished when the values of the structural constraints are at most 1. For an in-depth explanation of this problem, the reader is advised to read  \cite{Ferreira2019}.

Each individual is defined by an array composed of 22 variables, whose descriptions and domains can be seen in Table \ref{tab:dvs}. These variables are then utilized to calculate the price of the bridge and the respective values for the structural constraints. Given that for a bridge to be considered secure, all of the constraints need to be less or equal to 1, we resort to using the maximum value of these constraints instead of the individual values. In addition to the variable parameters, the bridges also have fixed ones, which are presented in Table. \ref{tab:fixed_dvs}.

In our experiments, we compare our results to a baseline solution, a bridge configuration optimized by the same approach used in Ferreira's et al. work \cite{Ferreira2019} with the same fixed parameters. This bridge has a cost of $91.354$ k\euro{} and a maximum of structural constraints of 0.9962.

\begin{table}
    \centering
    \caption{Values for the fixed parameters.}
    \begin{tabular}{lll}
        \hline
        Bridge Length (LTotal) & \hspace{0.5cm} & 220 meters \\
        Bridge Width &  & 4 meters \\
        Tower Height below deck &  & 10 meters \\
        \hline
    \end{tabular}
    \label{tab:fixed_dvs}
\end{table}

\begin{table}
    \centering
    \footnotesize
    \caption{Cable-stayed bridges design variables description and domain values.}\label{tab:dvs}
    \begin{tabular}{l|l|c}
        \hline
         \textbf{Variable Type} & \textbf{Description} & \textbf{Domain Values} \\
        \hline
        Discrete & \\
        \hline
        DV0    & Number of cables & 3,4,5,6,7\\
        \hline
        Geometry & &\\
        \hline
        DV1    & Central span (tower to tower distance) of the structure & $[0.9, 1.2]$\\
        DV2    & Distance between the first and second cables anchorage & $[0.7, 1.3]$\\
               & in the lateral span of the deck & \\
        DV3    & Distance between the tower and the first cable in the & $[0.7, 1.3]$\\
               & central span & \\
        DV4    & Distance between the last cable anchorage and the & $[0.7, 1.3]$\\
               & bridge symmetry axis & \\
        DV5    & Height of the towers & $[0.1, 2.0]$\\
        DV6    & Distance where the cables are distributed in the & $[0.1, 4.0]$\\
               & top of the towers & \\
        DV7    & Distance between the top of each tower & $[0.1, 1.3]$ \\
        DV8    & Distance between each tower at the base & $[0.1, 1.13]$ \\
        \hline
        Control & &\\
        \hline
        DV9    & Transversal stiffness of the tower-deck connection  & $[0.001, 1000]$  \\
        DV10   & Vertical stiffness of the tower-deck connection  & $[0.001, 1000]$  \\
        DV11   & Transversal damping of the tower-deck connection & $[0.001, 1000]$ \\
        DV12   & Vertical damping of the tower-deck connection & $[0.001, 1000]$ \\
        \hline
        Sectional and tensioning&\\
        \hline
        DV13   & Added mass of the concrete slab & $[0.1, 7.0]$ \\
        DV14   & Deck section  & $[0.1, 80.0]$  \\
        DV15   & Deck section (triangular section) & $[0.5, 1.3]$ \\
        DV16   & Tower section (rectangular hollow section)  & $[0.4, 1.5]$ \\
        DV17   & Tower section (rectangular hollow section)  & $[0.1, 20.0]$ \\
        DV18   & Tower section (rectangular hollow section)  & $[0.3, 20.0]$ \\
        DV19   & Tower section (rectangular hollow section)  & $[0.3, 9.0]$ \\
        DV20   & Cables pre-stress & $[0.7, 3.0]$ \\
        DV21   & Cables cross section & $[0.5, 9.0]$ \\
        \hline
    \end{tabular}
\end{table}


\section{The Approach}
To have a fair comparison with the previous works, we replicated the experiments conducted with the GA. The parameters used for the algorithm are almost the same as in \cite{DBLP:conf/gecco/CorreiaFM20}, apart from the number of generations, and can be seen in Table \ref{tab:Algs_params}.

The novelty that we add to this problem with this work is by experimenting with CMA-ES, implemented with Distributed Evolutionary Algorithms in Python (DEAP) framework \cite{DEAP_JMLR2012}. The parameters used can be seen in Table \ref{tab:Algs_params}. Since the sizes of the population are different, we ensure that the same number of evaluations is used, so that the results can be compared.

The initial individuals used to start the evolutionary process are created by uniformly sampling the domain intervals of each variable. This idea is also utilized in the mutation operator used in the Genetic Algorithm, meaning that when a gene is chosen to be mutated, the new value is also uniformly sampled from the domain of the specific variable. In order to deal with the unfeasible solutions generated by CMA-ES, we correct the specific variable to the minimum value of the domain if it is lower than it or to the maximum if it is greater than the maximum value. This process is not performed in the GA, because the variation operators ensure that the values of the variables are within the required domains, given that the crossover does not alter the values and the mutation operator, as previously stated, samples the new value from the domain. In CMA-ES, this correction is necessary because it is not guaranteed that the generated values are within the domain boundaries.

The fitness function used for both algorithms is presented in Eq. \ref{eq:fitness}.
$C(x)$ and $S(x)$ are the cost and the structural constraint value of individual $x$, respectively. The price returned by $C(x)$ is based on pre-determined pricing of the materials, and $S(x)$ returns the maximum value of the structural constraints of the individual~\cite{Ferreira2019}. In practice, we need to guarantee that the returned value of $S(x)$ is at most 1.0.

\begin{equation}
    \label{eq:fitness}
    f(x) = 
    \begin{cases}
    c_{r}/C(x), & \mbox{if } C(x) > c_{r} \\
    1 + 1 / S(x), & \mbox{if } C(x) < c_{r} \land S(x) > 1.0 \\
    2 - ( 1.0 -  S(x) ) + c_{r}/C(x), & \mbox{if } C(x) < c_{r} \land S(x) \leq 1.0 \\
    \end{cases}
\end{equation}

The fitness function aims to guide the population towards individuals that have a structural constraint of at most 1.0 and the lowest cost possible. First, by reducing the cost to a more acceptable value ($c_{r}$, it is fixed in our experiments, see Table \ref{tab:Algs_params}, but can be changed), then search for individuals that are feasible structurally by rewarding individuals that have $S(x)$ values closer to 1.0, and finally find individuals that are both feasible and cost-effective (we want the lowest cost possible). Although we aim to minimize the cost of the structures, it is important to notice that we want to maximize the fitness value, thus defining this problem as a maximization problem. In the first branch, the fitness ranges from 0 to 1, in the second, from 1 to 2, and in the third, it is greater than 2.

\begin{table}[H]
    \centering
    \footnotesize
    \caption{Algorithm's Parameters.}\label{tab:Algs_params}
    \begin{tabular}{lc}
        \hline\hline
        \textbf{Parameter}                   & \textbf{Value} \\
        \hline
        GA                          & ~ \\
        \hline
        Generations       & 40 000 \\
        Population size             & 10 \\
        Tournament size             & 3 \\
        Crossover operator          & Uniform Crossover \\
        Crossover rate (per gene)   & 0.5 \\
        Mutation operator           & per gene replacement \\
        Mutation rate per gene      & 0.1 \\
        \hline
        CMA-ES                      & Value \\
        \hline
        Generations                 & 8 000 \\
        $\mu$                       & 25 \\ 
        $\lambda$                   & 50 \\%
        $\sigma$                    & 0.5 \\
        \hline
        Common  & Value \\
        \hline
        Number of Runs & 30 \\
        Elite size & 1 \\
        $c_r$ fitness constant  & 150\\
        Number of evaluations & 400 000 \\
        \hline\hline
    \end{tabular}
\end{table}

\section{Experimental Results}

\begin{figure}
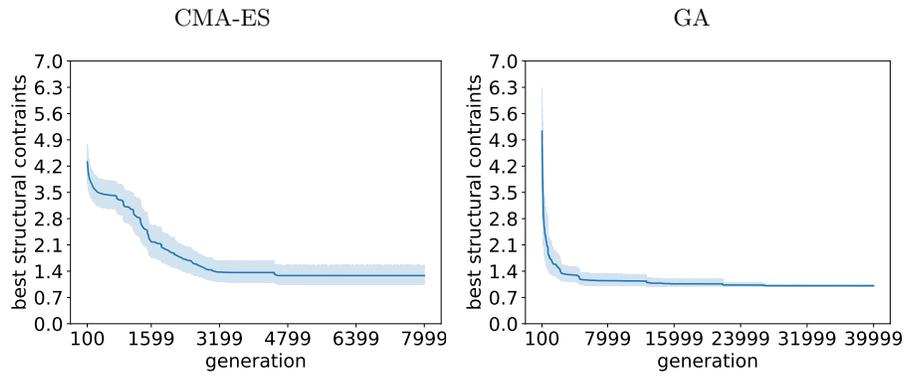

    \centering
    \begin{tabular}{c c}
        CMA-ES & GA\\
        \includegraphics[page=12, scale = 0.39, trim = {0cm 0cm 0.5cm 0cm}]{figures/CMA-ES.pdf} & 
        \includegraphics[page=12, scale = 0.39, trim = {0.5cm 0cm 0cm 0cm}]{figures/GA.pdf}
    \end{tabular}
    \caption{ Mean structural constraint values of the best individuals of 30 runs for CMA-ES (on the left) and for the GA (on the right) starting from generation 100 until generation 7 999 and 39 999 generations, respectively. We can see that CMA-ES stabilizes around the 4 800 generations mark and 28 000 for the GA. The first 100 generations ($[0 : 100[$) were not plotted due to the fact that the values of cost and structural constraints in these generations are extremely large, making the plots unreadable.}
    \label{fig:maxjjj_over_generations}
\end{figure}

\begin{figure}
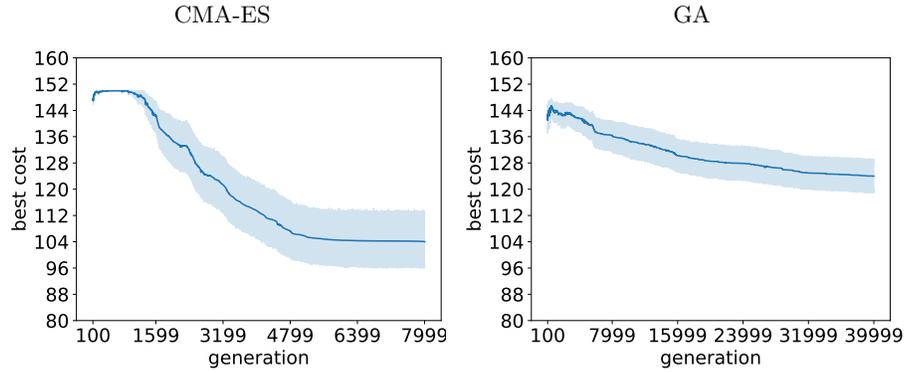

    \centering
    \begin{tabular}{c c}
        CMA-ES & GA\\
        \includegraphics[page=10, scale = 0.39, trim = {0cm 0cm 0.5cm 0cm}]{figures/CMA-ES.pdf} & 
        \includegraphics[page=10, scale = 0.39, trim = {0.5cm 0cm 0cm 0cm}]{figures/GA.pdf}
    \end{tabular}
    \caption{ Mean cost values of the best individuals of 30 runs for CMA-ES (on the left) and for the GA (on the right) starting from generation 100 until generation 7 999 and 39 999 generations, respectively. We can see that CMA-Es is able to achieve lower values of cost, however, it presents a higher variability between runs. It also appears to be stabilizing (we address this topic in the Experimental Results section). It appears that the GA is the opposite, continuing to optimize the cost even after the 40 000 generations. The explanation of why the first 100 generations are not plotted is presented in the caption of Fig.\ref{fig:maxjjj_over_generations}.}
    \label{fig:cost_over_generations}
\end{figure}

\begin{figure}
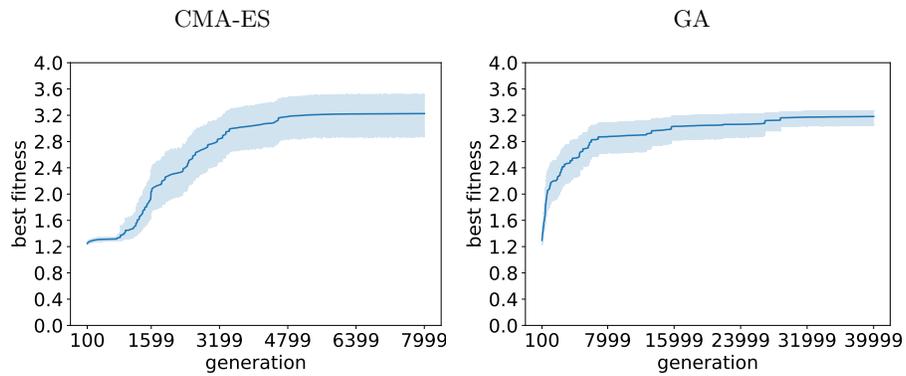

    \centering
    \begin{tabular}{c c}
        CMA-ES & GA\\
        \includegraphics[page=8, scale = 0.39, trim = {0cm 0cm 0.5cm 0cm}]{figures/CMA-ES.pdf} & 
        \includegraphics[page=8, scale = 0.39, trim = {0.5cm 0cm 0cm 0cm}]{figures/GA.pdf}
    \end{tabular}
    \caption{ Mean fitness values of the best individuals of 30 runs for CMA-ES (on the left) and for the GA (on the right) starting from generation 100 until generation 7 999 and 39 999 generations, respectively. It can be seen that CMA-ES is able to reach higher values of fitness, however, there is more variability between runs. The GA seems to be a more consistent algorithm, not showing as much variability. The explanation of why the first 100 generations are not plotted is presented in the caption of Fig.\ref{fig:maxjjj_over_generations}.}
    \label{fig:fitness_over_generations}
\end{figure}

The performance of the best individuals during the evolutionary process for the structural constraints $S(x)$ are presented in Fig. \ref{fig:maxjjj_over_generations}. Fig.~\ref{fig:cost_over_generations} presents the results regarding the cost $C(x)$, whilst Fig.~\ref{fig:fitness_over_generations} presents the results for the fitness function, $f(x)$. Results are averages of 30 independent runs.

\begin{figure}
    \centering
    \includegraphics[page=5, scale = 0.8]{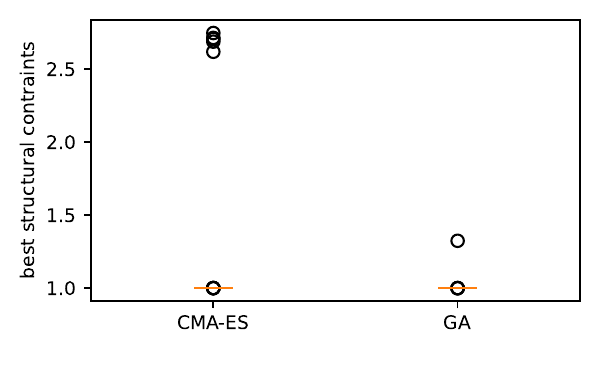}
    \caption{Boxplot of the structural constraint of the best individual of each run (30 in total) for CMA-ES (on the left) and the GA (on the right).}
    \label{fig:boxplot_maxjjj}
\end{figure}

Looking at Fig.\ref{fig:maxjjj_over_generations}, one can see that both approaches gradually improve the values of the structural constraints, by gradually getting close to the upper limit value of 1. Looking at the right panel, one can see that the GA has a steep descent in the value of the structural constraint, whilst CMA-ES performs a more slow descent. By the end of the optimization process, both approaches have reached approximately the same structural constraint values. To better understand the differences between the approaches we presented a boxplot of the results regarding the structural constraint for the best individuals. Looking at Fig.~\ref{fig:boxplot_maxjjj} one can see that CMA-ES is capable of optimizing this objective, but it has some runs where it fails by a relatively bigger margin, leading to a large mean value. The GA also fails (1 seed out of the 30), but by less. Both curves converge close to 1.0, which is the desired behavior, meaning that the structural safety of the bridge is being optimized.

In what concerns the cost, the results are depicted in Fig.~\ref{fig:cost_over_generations}. Looking at the results, one can see that during the first 1500 generations, the CMA-ES (left panel) does not seem to improve the values of the initial solutions. In fact, looking at the graph, one can see that it slightly increases cost. However, after generation 1500, the approach starts to rapidly improve the cost of the bridge. The curve stabilizes around generation 6000, which might be an indication that the CMA-ES reaches an optimum. The GA (right panel) exhibits roughly the same behavior in what concerns the optimization trend. These results might be explained by the fact that both approaches, in the first generations, focus on obtained bridges that have a good value in terms of structural constraints. After having such bridges the approaches start to reduce the cost, without compromising the integrity and safety of the bridge.

\begin{figure}
    \centering
    \includegraphics[page=3, scale = 0.8]{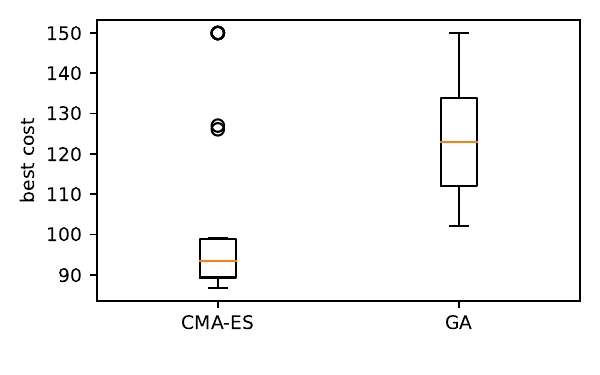}
    \caption{Boxplot of the cost of the best individual of each run (30 in total) for CMA-ES (on the left) and the GA (on the right).}
    \label{fig:boxplot_cost}
\end{figure}

Another interesting result is that by the end of the evolutionary process, the best solutions obtained by the CMA-ES have a much lower cost than the ones discovered by the GA. To help with this analysis, we created a boxplot of the cost values for both approaches and show them in Fig.~\ref{fig:boxplot_cost}. Whilst CMA-ES is not as good as the GA at optimizing the structural constraints, it reaches brilliant results in terms of cost. In fact, one can see that the CMA-ES approach can not only find bridges with lower costs but also finds them consistently given the lower variance obtained when compared to the GA.

\begin{figure}
    \centering
    \includegraphics[page=1, scale = 0.8]{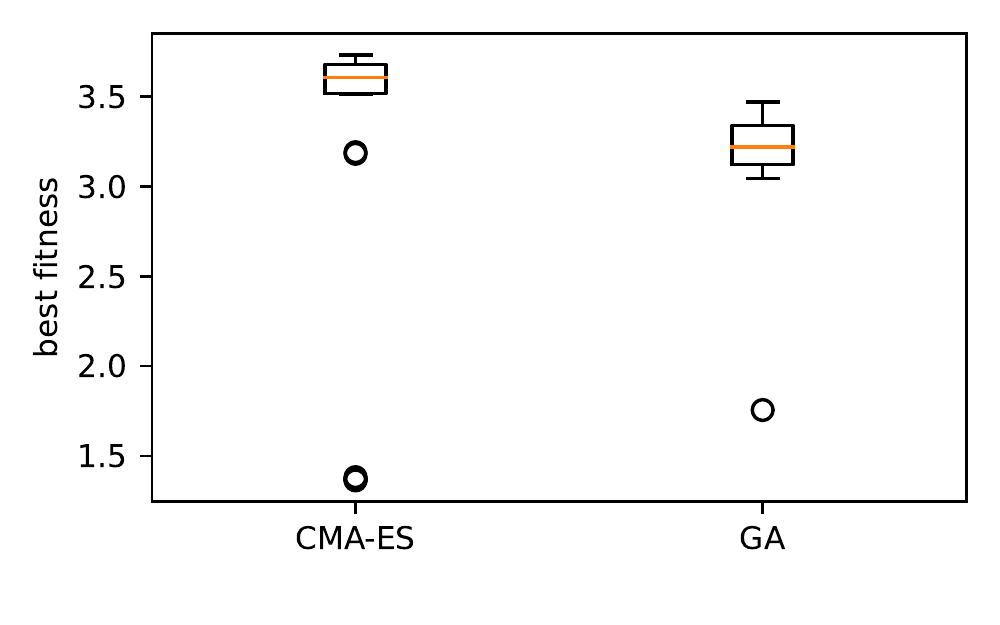}
    \caption{Boxplot of the fitness of the best individual of each run (30 in total) for CMA-ES (on the left) and the GA (on the right).}
    \label{fig:boxplot_fitness}
\end{figure}

We also show the fitness plots, Fig.\ref{fig:fitness_over_generations} and \ref{fig:boxplot_fitness}, to show the results for the $f(x)$ that combines both the cost $C(x)$ and the structural constraint $S(x)$. 

Finally, Table \ref{tab:stats_experiments} summarises the results.  Looking at the values, it seems that CMA-ES, on average, does not differ that much from the GA. Even though  the curves for the structural constraints are similar and the curves of the cost are very different, it appears that our fitness function is not very good at stretching the fitness values when the structural constraints are already satisfied, leading to similar values (but different) of fitness for candidate solutions that have a relatively different cost. For the optimization itself, it still classifies a better individual with higher fitness, however, when plotted, the difference is not very evident.

To understand if there are meaningful differences between the two approaches, we performed a statistical analysis. Since the samples do not follow a normal distribution, we used the Mann-Whitney non-parametric test with a significance level of $\alpha=0.05$. The effect sizes are presented in Table \ref{tab:statistic_tests}, and it can be observed that there is a large effect size in all the metrics, meaning that the differences between both approaches are significant.


\begin{table}
    \centering
    \footnotesize
    \caption{Results of the statistical analysis using the Mann Whitney U test with a significance level $\alpha=0.05$.}\label{tab:statistic_tests}
    \begin{tabular}{l c}
        Feature & Effect Size\\
        \hline
        fitness & -0.510 \\
        $C(x)$ & 0.510 \\
        $S(x)$ & -0.702 \\
        \hline
    \end{tabular}
\end{table}

Table \ref{tab:top} presents the cost and the value of the structural constraint of the best solution for every approach. CMA-ES was able to achieve multiple solutions that beat the baseline (see Improvement Rate in Table \ref{tab:statistic_tests}) (with a good level of diversity, because the solutions have different numbers of cables), while the GA could not do it once (see Table \ref{tab:stats_experiments}). With the help of the differences, one can see that both the approaches optimized the structural constraints, however, CMA-ES was able to reduce the cost of the bridge by more than $4$ k\euro{}, which is a substantial amount. 

\begin{figure}
    \centering
	\begin{subfigure}
		\centering
	    \includegraphics[page=1, scale = 0.6, trim = {2cm 10cm 2cm 11cm}, clip]{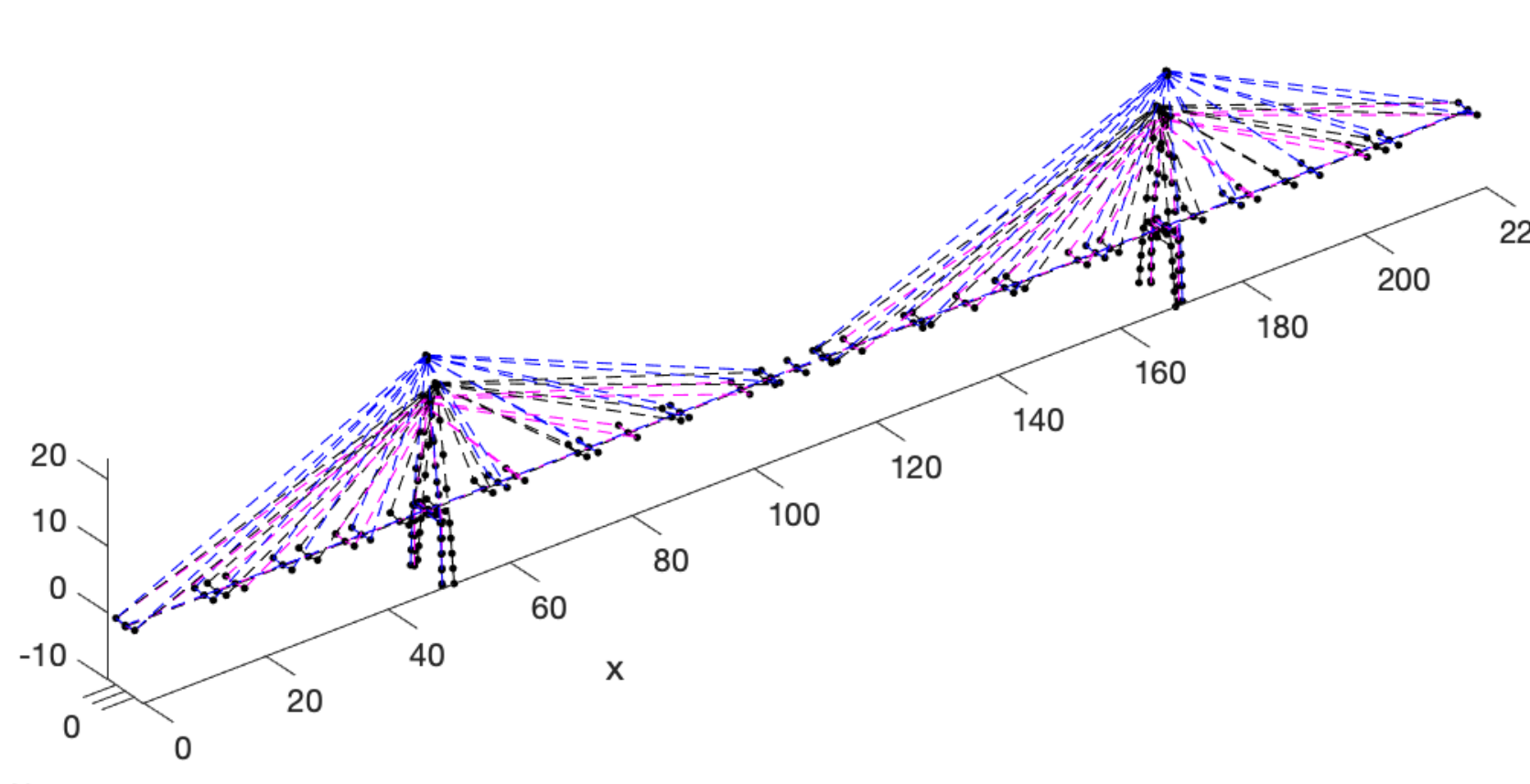}
	\end{subfigure}
	\vfill
	\begin{subfigure}
		\centering
		\includegraphics[page=1, scale = 0.6, trim = {2cm 12cm 2cm 12cm}, clip]{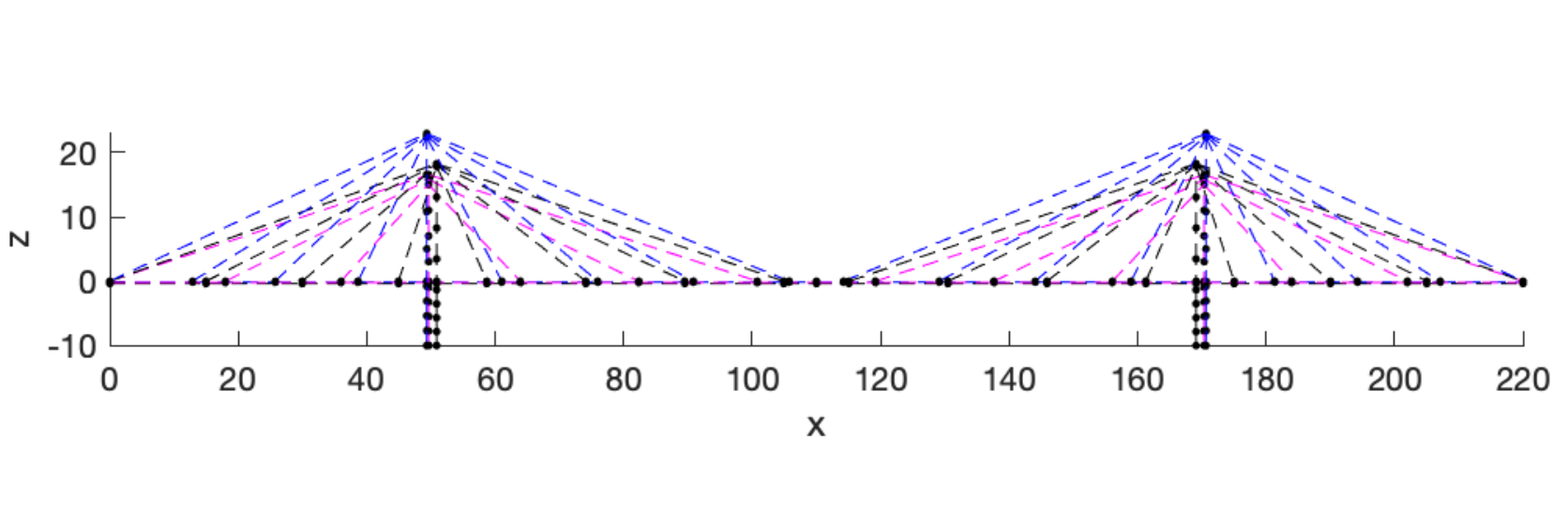}
	\end{subfigure}
    \caption{Baseline bridge (black), versus the bridges optimized by the GA (pink) and by CMA-ES
    (blue).}
    \label{fig:brigde_baseline_cmaes_ga}
\end{figure}

Fig.\ref{fig:brigde_baseline_cmaes_ga} shows how the best bridges evolved by the two algorithms compare to the baseline and to each other. We can see that the CMA-ES is visually distinct from the baseline solution due to its height, but the one evolved by the GA is significantly different from the rest because it uses 3 cables, while the others use 4.

\begin{table}
  \centering
  \caption{Best solution of every approach and the difference between cost and structural constraints against the baseline approach. The best values for both the cost and the structural constraints are in bold.}
  \begin{adjustbox}{max width=\textwidth}
    \begin{tabular}{cccccc}
          & Baseline(B) & diff(B, GA) & best GA & diff(B, CMA-ES) & best CMA-ES \\
    \hline
    $C(x)$ & 91.354 & -10.840 & 102.194 & 4.656 & \textbf{86.698} \\
    $S(x)$ & \textbf{0.996} & -0.004 & 1.000 & -0.004 & 1.000 \\
    \hline
    \end{tabular}%
  \end{adjustbox}
  \label{tab:top}%
\end{table}%

CMA-ES finds really good solutions to this problem but appears to be a little extreme, meaning that when it is able to find a good solution it is really good, but when it is not able to, the result is not satisfactory, being expensive and not even structurally safe.

Table \ref{tab:stats_experiments} further cements what was previously said, showing that, on average, CMA-ES is worse in optimizing $S(x)$ but is able to greatly reduce the cost of the structures when compared to the GA. However, the GA is more consistent, presenting less variability, seen by the standard deviation values.

\begin{table}
    \centering
    \footnotesize
    \caption{Stats from the experiments. Mean(...) is the average of the 30 runs and Improvement Rate is the rate of runs that were able to beat the baseline. The average values are presented along with the respective standard deviation.}\label{tab:stats_experiments}
    \begin{tabular}{l| c c c c}
        ~ & Mean(fitness) & Mean($C(x)$) & Mean($S(x)$) & Improvement Rate \\
        \hline
        CMA-ES & 3.225 ($\pm$ 0.862)  & 104.005 ($\pm$ 23.042) & 1.282 ($\pm$ 0.652) & 11/30 \\
        GA & 3.181 ($\pm$ 0.296) & 123.987 ($\pm$ 13.124) & 1.01 ($\pm $0.059) & 0/30 \\
        \hline
    \end{tabular}
\end{table}

\begin{figure}
    \centering
    \includegraphics[page=1, scale = 0.8]{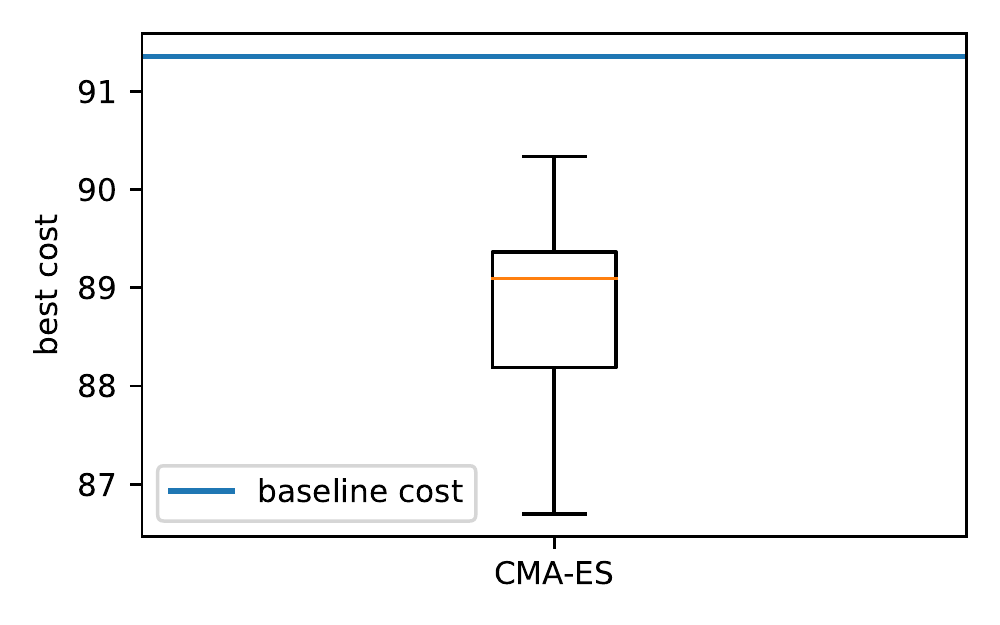}
    \caption{Boxplot of the cost of the best 11 seeds of CMA-ES plotted with the baseline cost. This better highlights how much the CMA-ES was able to beat the baseline. All of the solutions that beat the baseline have a structural constraints value of at most 1.0, so these values are not plotted here because they would be basically all the same.}
    \label{fig:boxplot_cost_11_best_CMA-ES}
\end{figure}

We decided to include Fig.\ref{fig:boxplot_cost_11_best_CMA-ES} because we wanted to show how much CMA-ES improved the cost in relation to the one of the baseline. For this image, only the results of the 11 seeds that beat the baseline were included. We do not present a figure for the structural constraints, because, as previously stated, we are only using the results of the seeds that beat the baseline, meaning that the values of the structural constraint are all at most 1.0, given the inverse nature between the cost and the structural constraint.

\section{Conclusions}






The design of Cable-stayed bridges (CSB) is one of the most complex designs in bridge engineering since they are highly static indeterminate and cannot be calculated by hand in a short amount of time. This task is mostly handled manually by Civil Engineers, where most of the research on CSB employs gradient-based optimization techniques which require programming the sensitivities of the problem. In this work, we perform a comparison of the performance of two evolutionary approaches, a standard GA and CMA-ES, in terms of cost and structural constraints. We further complement this analysis by comparing the results of both with a previously gradient-based optimized solution found in the literature.

In our results, CMA-ES was able to achieve a cost value of 86.698 k\euro{}, beating the baseline and GA costs, 91.354 k\euro{} and 102.194 k\euro{} respectively, while maintaining the structural constraints in acceptable values according to the safety codes. The behaviour of the GA and CMA-ES approach was analyzed in 30 different seeded runs. Under the same budget of evaluations, the GA was not able to beat the baseline solution not even once in terms of cost, despite being more consistent at optimizing the structural constraints when compared with the CMA-ES. The CMA-ES was able to beat the baseline 11 times by a significant margin. Statistical tests were performed with the results of the 30 runs of each algorithm, and the differences between the two were significantly different. The results suggest that CMA-ES performs better, under this setup, for this problem than a standard GA. 


In future work, we intend to use quality-diversity algorithms, optimizing both the objective and exploring distinct solutions, to expand our knowledge of the search space as well as being able to retrieve multiple high-performing solutions from a single run and avoid local optimums that may exist.




\section*{Acknowledgements}
This research was partially funded by the project grant BEIS (Bridge Engineering Information System), supported by Operational Programme for Competitiveness and Internationalisation (COMPETE 2020), under the PORTUGAL 2020 Partnership Agreement, through the European Regional Development Fund (ERDF) and by the FCT - Foundation for Science and Technology, I.P./MCTES through national funds (PIDDAC), within the scope of CISUC R\&D Unit - UIDB/00326/2020 or project code UIDP/00326/2020*

\bibliographystyle{splncs04}
\bibliography{fernandes} 
\end{document}